# Multi-granularity Association Learning Framework for on-the-fly Fine-Grained Sketch-based Image Retrieval

Dawei Dai[1], Xiaoyu Tang[1], Shuyin Xia[1*], Yingge Liu[1], Guoyin Wang[1], Zizhong Chen[2]
1. College of Computer Science and Technology, Chongqing University of Posts and Telecommunications, Chongqing, China
2. the department of Computer Science and Engineering, University of California, Riverside

**Abstract**

Fine-grained sketch-based image retrieval (FG-SBIR) addresses the problem of retrieving a particular photo in a given query sketch. However, its widespread applicability is limited by the fact that it is difficult to draw a complete sketch for most people, and the drawing process often takes time. In this study, we aim to retrieve the target photo with the least number of strokes possible (incomplete sketch), named on-the-fly FG-SBIR (Bhunia et al. 2020), which starts retrieving at each stroke as soon as the drawing begins. We consider that there is a significant correlation among these incomplete sketches in the sketch drawing episode of each photo. To learn more efficient joint embedding space shared between the photo and its incomplete sketches, we propose a multi-granularity association learning **framework** that further optimizes the embedding space of all incomplete sketches. Specifically, based on the integrity of the sketch, we can divide a complete sketch episode into several stages, each of which corresponds to a simple linear mapping layer. Moreover, our framework guides the vector space representation of the current sketch to approximate that of its later sketches to realize the retrieval performance of the sketch with fewer strokes to approach that of the sketch with more strokes. In the experiments, we proposed more realistic challenges, and our method achieved superior early retrieval efficiency over the state-of-the-art methods and alternative baselines on two publicly available fine-grained sketch retrieval datasets.

## Introduction

Recently, considering the rapid proliferation of various electronic touch screen devices, more convenient hand-painted input conditions have been provided for the majority of users. This makes hand-painted graphics (sketches) increasingly popular in people's daily lives, works, and entertainment, and thus more and more hand-painted data appear on the Internet. Consequently, sketch-related problems have become a topic of considerable research interest in the field of computer vision (Collomosse et al. 2019; Dey et al. 2019). Considering these fields, fine-grained sketch-based image retrieval (FG-SBIR) (Yu et al. 2016; Song et al. 2018; Pang et al. 2019; Sain et al. 2020) has received particular attention because of its potential commercial applications, which address the problem of retrieving a particular photo for a given query sketch, and the query sketch was usually complete. Although there has been great progress in the field of FG-SBIR over the years, two key points hinder its wide application practically. First, a sketch can convey fine-grained details more easily than text/tag; however, drawing a sketch takes time and is often slower than clicking a tag for typing the keyword practically. Second, drawing a complete sketch requires skills, and the drawing skills of different users vary greatly.

A new FG-SBIR problem (Bhunia et al. 2020) that assumes an on-the-fly setting was proposed to overcome the aforementioned barriers. The retrieval was performed at each stroke drawn or fixed time interval. Such an on-the-fly FG-SBIR aims to retrieve the target photo by a incomplete sketch with as few strokes as possible. On-the-fly FG-SBIR is essentially a cross-modal matching problem, which is typically solved by learning a joint embedding space, where the feature vector is shared between target photo and their sketch modalities. Consequently, learning the efficient joint embedding space of **incomplete or poorly** drawn sketches and their corresponding target photo is the key point to resolve the on-the-fly FG-SBIR problem.

A popular framework uses a triplet network to learn the joint embedding space for the FG-SBIR problem. However, for the problem of on-the-fly FG-SBIR, one photo can create multiple incomplete sketches, and such diversity can obviously confuse the triplet network. In fact, incomplete sketches of one photo are not independent of each other, a great correlation exists between these sketches in a sketch-drawing episode. Inspired of this, in this study, we propose a framework of multi-granularity association learning to deal with the diversity and correlation of the incomplete sketches. We first divide a complete sketch episode into several stages according to the integrity (the number of strokes) of a sketch, and design an independent learnable mapping layer for each stage, between which we perform the association learning for the current incomplete sketches to their later sketches. We applied our method to two publicly available fine-grained sketch retrieval datasets and achieved superior early retrieval efficiency over the state-of-the-art methods. As shown **Fig.1**, we show one example of our methods compared to SOTA result (Bhunia et al. 2020); We observe that the target photo can be retrieved in the top-5 list using fewer strokes than that of the baseline model; Therefore, practically, users can stop as early as their target photos appear in the result list.

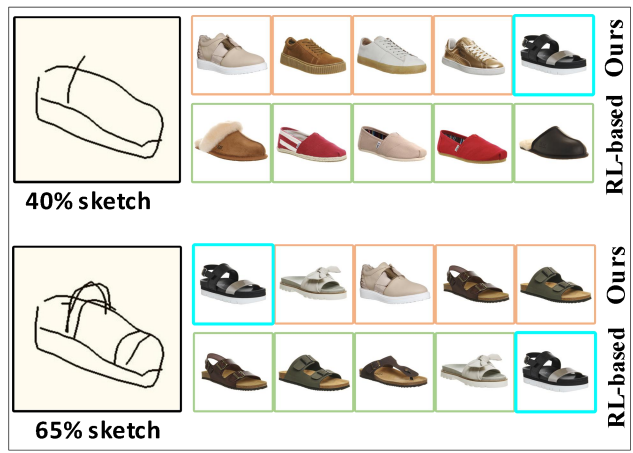

**Figure 1**. Demonstration of our methods' ability to retrieve (top-5 list) the target photo using fewer number of strokes than the baseline method (Bhunia et al. 2020).

Our contributions can be summarized as follows. (a) Considering the on-the-fly FG-SBIR framework, we develop a multi-stage model for the sketches with different completeness to avoid the diversity of incomplete sketches confusing the model. (b) We proposed a framework of multi-granuarity association learning for progressive incomplete sketches, so that the embedding space of each incomplete sketch approximates that of both subsequent sketch and its corresponding target photo, and do so with the minimum possible drawing. (c) We proposed several new realistic challenges for the on-the-fly FG-SBIR problem, and extensive experiments on two public datasets demonstrate the superiority of our proposed on the problem of on-the-fly FG-SBIR.

## Related Work

**Category-level SBIR:** Sketch-based image retrieval is considered as a cross-modal matching problem that is typically solved by learning a joint embedding space in which the semantic content shared between the photo and sketch modalities is preserved. This type of retrieval task aims to retrieve target photos at the **category level** in a given query sketch, where if the retrieved photo and query sketch belong to the same category, the retrieval is considered successful. Therefore, the category-level SBIR problem focuses more on the differences between **categories**. Previous methods used manual descriptors to construct global or local joint representations shared between photos and their associated sketches (Qi et al. 2015; Tolias and Chum 2017). These include SIFT (Lowe et al. 1999), HOG (Hu and Collomosse 2013), edge local direction histograms (Saavedra et al. 2014), and learning key shapes (Saavedra et al. 2015). Recently, deep neural network models have been used to extract more effective sketch features (Qian et al. 2017; Cong et al. 2020). For example, Yu et al. (2015) proposed a multi-scale channel neural network that optimized the sketch based on the feature information in the sketch. For the first time, the recognition performance on a large-scale sketch benchmark dataset was 1.8% higher than that of humans. Subsequently, classical ranking losses, such as contrast or triplet loss (Yu et al. 2016), have been used in SBIR problems to narrow the embedding space of sketch-photo pairs. To deal with the problem of large-scale image retrieval, researchers have highlighted many methods for constructing a hash index around the basic idea of a hash. For example, Liu et al. (2017) proposed a semi-heterogeneous deep architecture (deep sketch hashing), which aimed to reduce the computational cost of retrieval, including retrieval time and memory occupation, Song et al. (2019) proposed an edge-guided cross-domain learning method based on shape regression to address the problem of mapping the sketch and image domains to the common space domain.

**Fine-grained SBIR:** Fine-grained SBIR is a recent addition to the SBIR field and has been less studied than the category-level SBIR task; it retrieves photos at the **instance level**. In contrast to category-level SBIR, each instance can be regarded as a category. Therefore, the FG-SBIR is clearly more challenging than that of Category-level SBIR. Yu et al. (2016) constructed a neural model based on a pre-trained model using a triple ranking loss and combined the edge image and sketch into a mapping pair, which improved the efficacy of spatial mapping. Furthermore, Zhang et al. (2017) realized end-to-end image retrieval using heterogeneous networks to deal with the problem of information loss when extracting features from edge images. Wang et al. (2019) described fine-grained sketch retrieval as a coarse-to-fine process and proposed a deep cascaded cross-modal ranking model that exploits all the beneficial multi-modal information in sketches and annotated images. Peng et al. (2019) proposed a novel unsupervised learning approach to model a universal manifold of prototypical visual sketch traits as a domain generalization problem to identify cross-category generalization for FG-SBIR. Zhu et al. (2019) designed a gradually focused bi-linear attention model to extract detailed information more effectively. Sain et al. (2020; 2021) proposed a cross-modal network method with hierarchical co-attention, which matches the sketch and photo at corresponding hierarchical levels, and then uses cross-modal VAE and meta-learning to realize style agnostic SBIR. Du et al. (2020) proposed a progressive training strategy effectively to integrate features with different granularity. In the latest research by Choi et al. (2019), Huang F et al. (2019), Huang Z et al. (2019), Collomosse et al. (2019), and Bhunia et al. (2020), an interactive framework was proposed to relieve sketch retrieval from being limited to static search.

**Incomplete Sketch:** Studies on sketch-based image retrieval (Hongyu et al. 2019; Kamyar et al. 2019; Li et al. 2020; Xie et al. 2021) have mainly focused on sketch synthesis and completion. The efficacy of both the attention mechanism (Yu et al. 2018) and VAE (Zheng et al. 2019) has been verified for sketch completion. Ha and Eck (2017) proposed a sketch-RNN to generate sketches automatically of specified types based on pen strokes. Moreover, Chen et al. (2017) used a CNN to replace the bidirectional recurrent neural network encoder, which performed better than the latter in generating multiple types of sketches. Cao et al. (2019) proposed a CNN-based autoencoder to capture pixel-level positional information between strokes to generate high-quality sketches and used condition vectors as distinguishers to improve the performance on multi-category sketches. Liu et al. (2019) proposed a method that uses conditional GANs to complete sketches at the image-to-image level to assist with recognition. Aksan et al. (2020) proposed a generative model for complex free-form structures based on stroke-based drawing tasks, which regards drawings as a collection of strokes that form complex structures (e.g., flow charts), and can simulate the appearance of individual strokes and the compositional structure of larger diagrammatic drawings. Lin et al. (2020) proposed a bidirectional encoder representation from a transformer (Sketch-BERT) model to retrieve images from sketches. Reinforcement learning-based models also worked well in learning stroke-wise representations from pixel images of sketches. For example, Zhewei et al. (2019) constructed a differentiable neural renderer to render strokes based on deep reinforcement learning to improve the quality of recreated images; Contrary to the aforementioned two-stage inference frameworks, Bhunia et al. (2020) proposed a new framework known as the on-the-fly FG-SBIR, in which the retrieval was conducted at every stroke drawn and a reinforcement learning-based method was used to optimize the embedding space of incomplete sketches, thereby improving on earlier retrieval performance.

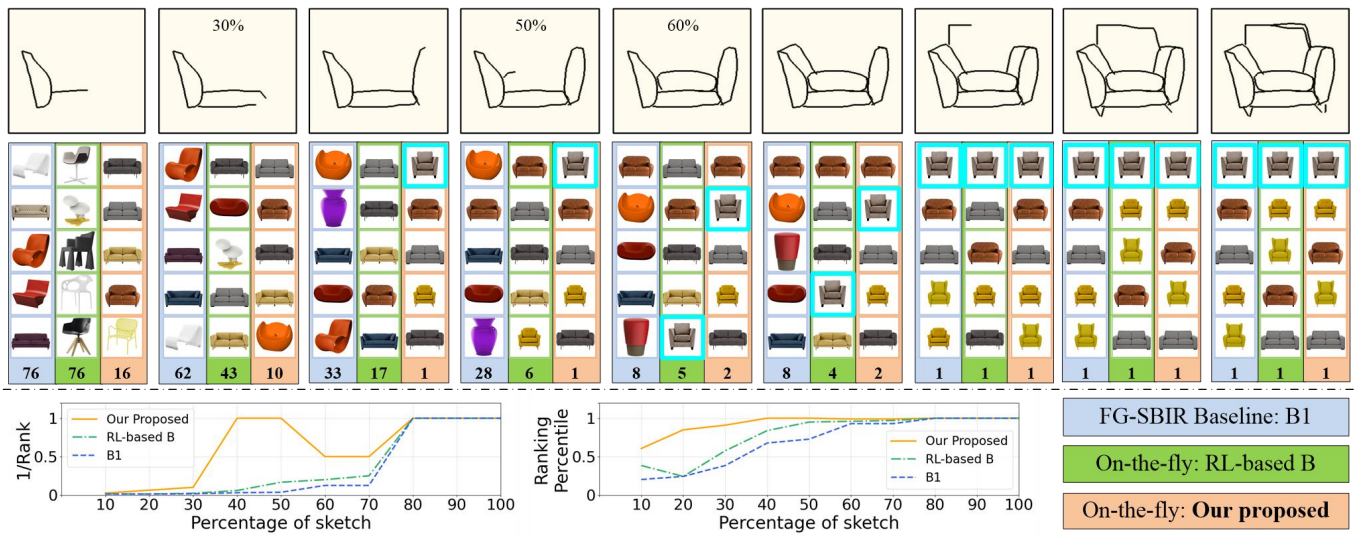

**Figure 2**. Illustration of our proposed for on-the-fly framework over two baselines. One is an FG-SBIR baseline (B1) (Szegedy et al. 2016) trained with completed sketches, and another is a reinforcement learning-based method (RL-based B) (Bhunia et al. 2020) trained with incomplete sketches. Considering this particular example, our method needs only 30% of the complete sketch to include the true match in the top-10 rank list, compared to 50% for RL-based B and 60% for B1. The top-5 photo images retrieved by the three methods are shown here. The number at the bottom denotes the paired (true match) photo's rank at every stage.

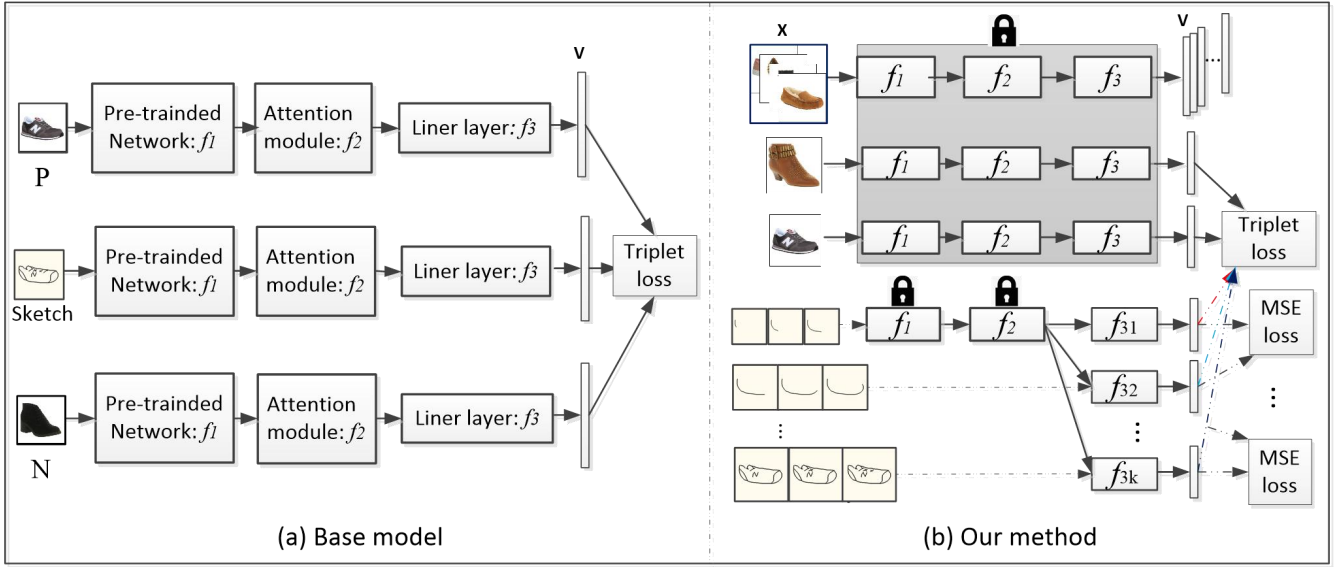

Figure 3. Overview of our approach. (a) Base model: a conventional FG-SBIR framework; (b) Our proposed on-the-fly FG-SBIR. The locks signify that the weights are fixed during learning.

## Methodology

### Overview

We proposed a new framework of Multi-granularity Association Learning (**MGAL**) to address on-the-fly FG-SBIR problem. Our study aims to learn efficient embedding spaces for the sketch drawing episode, and expects to retrieve the target photo at the earliest stroke possible (See **Fig.2**, an example of our proposed works in practice). An overview of our proposed is shown in **Fig.3**, where we first train a state-of-the-art FG-SBIR model (Szegedy et al. 2016; Bhunia et al. 2020) as the base model using a triplet loss; Thereafter, we maintain the photo branch and learn the sketch branch using differentiable layer over incomplete sketch. Here, several sketches with consecutive strokes correspond to one differentiable layer. Finally, a triplet loss and MSE loss were used for the output of each differentiable layer.

Formally, the base model learns an embedding function **F** that maps a complete sketch $s$ and its target photo $x$ to a D-dimensional feature vector $v$. We obtain a list of vectors $V = \{F(x_i)\}_{i=1,...,n}$, considering a given gallery of $n$ photos $X = \{x_i\}_{i=1,...,n}$. Regarding a given query of sketch $s$, we obtain its embedding vector using our method (see **Fig.3(b)**), and we obtain the top-k retrieved photos from **V** based on the pairwise distance metric. If the target photo first appears in the top-k list at the current stroke, we consider the top-k accuracy to be true for that sketch. Since, one photo can create a series of sketches, in which a sketch rendering operation (Bhunia et al. 2020) is used to produce rasterized sketches.

### Base Model

We first designed a neural model to learn the joint embedding space shared between the photo and its complete sketch. In order to verify the effectiveness of our framework, we use the neural network (Bhunia et al. 2020) in which there are three CNN model branches with shared weights corresponding to a positive photo, a query complete sketch, and negative photo as shown in **Fig. 3(a)**. The CNN model can be divided into three parts. The first part is a pre-trained neural model $f_1$, which is used to extract the feature for the input images (including complete sketch, positive photo, and negative photo). The InceptionNet (Szegedy et al. 2016) models trained on ImageNet are used as the pre-trained model as shown in **Eq. (1)**, where $x$ and $B$ indicate the input image (photo or sketch) and its corresponding feature maps. The second part $f_2$ furthers learning the embedding vector of the complete sketch and its target photo as shown **Eq. (2)**; two attention mechanisms ($f_{att}$) including spatial and channel (Dey et al. 2019) attention can be used. The third part $f_3$ was used to reduce the dimension of high-dimensional vectors ($V_H$) to obtain the low-dimensional vectors ($V_L$); we use a simple linear mapping (**A**) as shown in **Eq. (3)**. We design a triplet loss to learn the joint embedding space shared between photos and their sketches, and we consider only the complete sketch in the base model.

$$B = f_1(x) \quad (1)$$
$$V_H = f_2(B) = Global\_pooling(B + B \cdot f_{att}(B)) \quad (2)$$
$$V_L = A \cdot V_H \quad (3)$$

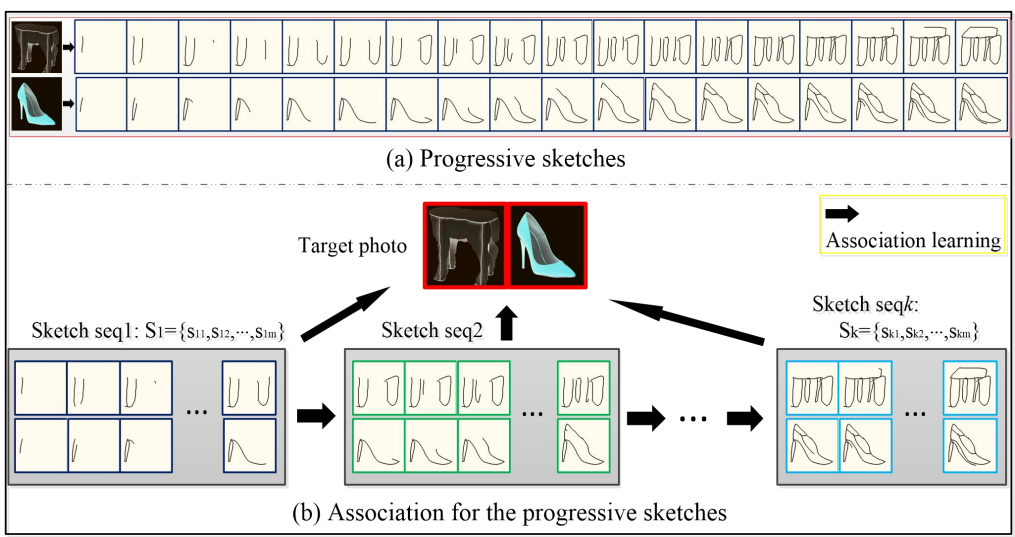

**Figure 4.** Overview of multi-granularity association learning framework. (a) Complete sketch drawing episode are progressive and highly correlated; (b) we divide a complete sketch drawing episode into several parts, where each sketch in one part is related to its subsequent sketches in the next part and corresponding target photo.

## Multi-Granularity Association Learning Framework
### Overview
Considering the problem of on-the-fly FG-SBIR, one photo corresponds to a series of incomplete sketches. As shown in **Fig.4(a)**, these incomplete sketches are obviously progressive, and there is a strong relationship between them. Inspired of this, we proposed a multi-granularity association learning for such progressive sketches (see **Fig.4(b)**). We can divide the complete sketch episode into $k$ parts according to the integrity of the sketch, each of which contains multiple progressive sketches that need to learn a mapping function $f_{3k}$ (see **Fig. 3(b)**). The embedding space of each incomplete sketch was guided to approach that of the subsequent sketches and target photo. We note that $k$ was a super parameter.

### Multi-stage Model
Owing to the creation of multiple incomplete sketches of each photo, a diversity that can confuse the base network is created, and our base model only learns the shared embedding space between the photo and its complete sketch. Considering a target photo, all of its incomplete sketches are one part of its complete sketch, and when there are many strokes in a sketch, there is a strong correlation between these incomplete sketches. This indicates that we can use the embedding space of the sketch with current strokes to associate it to that of the subsequent sketch by learning a mapping function (See function $f_{31}, f_{32}, f_{3k}$ in **Fig. 3(b)**), where we only use a simple linear mapping layer (See $A_1, A_2, ..., A_k$ in **Eq. (4)**).

$$f_{3i}(V_S) = A_i \cdot V_S, i = \{1, 2, ..., k\} \quad (4)$$

One sketch usually contains many strokes, and a single stroke is a small step for one sketch. Consequently, it is not necessary to learn a linear mapping for each single incomplete sketch, and several progressive sketches can share one linear mapping. The number of sketches that share a mapping layer can be a hyper-parameter. Considering our method, we split one complete sketch episode into several parts evenly. We can approximately determine which part every sketch belongs to by stroke number or designing a classifier to identify the sketch integrity.

**Multi-Granularity Association Learning**
Theoretically, one sketch with more strokes can contain more details or information than that with fewer strokes. And thus, the more complete the sketches are, the better is the retrieval performance. As mentioned in the previous section, sketches of one photo are strongly related to each other. Consequently, **the first goal of** our framework is to make the embedding space of each current incomplete sketch (embedding vector $v_{[i,j]}$) as close as possible to that of its subsequent sketches ($v_{[i+1,rnd]}$) as shown in **Eq. (5)** and **Fig.4(b), rnd** refers to a vector of a sketch that selected at random in the next module. Considering the $i_{th}$ sketch seq as an example, each sketch in $S_i$ is close to any one sketch in $S_{i+1}$ at each step of the learning. The embedding vector of the sketch is used to match the embedding vectors of all the photos in the database to calculate the distance and select a photo with the smallest distance to verify whether the target photo is among them. **The second goal** is to make the embedding space of an incomplete sketch as close as possible to that of its target photo ($v_p$). This can be performed through a triplet loss (see **Eq. (6)**), term of $v_n$ indicates non-target photo (negative sample).

$$J1_i(\theta) = \min \left( \sum_{j=0}^{m} d(v_{[i,j]}, v_{[i+1,rnd]}) \right) \quad (5)$$

$$J2_i(\theta) = \max \left( \sum_{j=0}^{m} (d(v_{[i,j]}, v_p) - d(v_{[i,j]}, v_n) + \alpha, 0) \right) \quad (6)$$

## Experiment

**Dataset**
We used QMUL-Shoe-V2 (Muhammad et al. 2018; Song et al. 2018; Pang et al. 2019) and QMUL-Chair-V2 (Song et al. 2018) datasets that were designed for FG-SBIR to train our base model, and their rasterized sketch images that had been specifically designed for the on-the-fly FG-SBIR problem were used to train our models and to evaluate their retrieval performance over different stages of a complete sketch drawing episode. QMUL-Shoe-V2 contained 6730 sketches and 2000 photos, of which 6051 and 1800, respectively, were used to train the models, and the rest were used to test our model. QMUL-Chair-V2 contained 2000 sketches and 400 photos, of which 1275 and 300 were used to train our model, and the rest were used to test our model.

## Implementation Details

We implemented our method in PyTorch following the practice of Bhunia (Bhunia et al. 2020) and conducted experiments on a 40 GB NVIDIA A100 GPU. We adopted the Inception-V3 (Szegedy et al. 2016) network pre-trained on ImageNet datasets (Russakovsky et al. 2015) as the backbone network to extract the features for both the photos and sketches. We used the Adam optimizer (Kingma and Ba 2014) with a mini-batch size of 16 (96 for Shoe-V2) and set $D = 64$ as the embedding space dimension. We rendered the sketch images at $T = 20$ steps. While training the backbone network, we used the triplet loss with a margin of 0.3 for 100 epochs at a learning rate of 0.0001. The channel of the extracted feature map for the sketch branch was 2048. Thereafter, we trained the final liner layer of the sketch branch for 500 epochs (keeping $f_1$ and $f_2$ fixed). The learning rate started from 0.001 and dropped to 0.0001 after 100 epochs. We used the triplet and MSE losses at a ratio of 1:1 with a weight decay of 0.0001.

## Evaluation Metric

Regarding the frame of the on-the-fly FG-SBIR, we prioritize the target photo appearing at the top of the list. Thus, the percentage of sketches with true match photos appearing in the top-q list (Acc.@q accuracy) was used to quantify the performance. Moreover, m@A (the ranking percentile) and m@B (1/rank versus percentage of sketch) (Bhunia et al. 2020) were used to capture the early retrieval performance for the incomplete sketches. Considering this context, a higher value of m@A and m@B indicates a better performance during the early sketch retrieval.

## Baseline Methods

The framework of on-the-fly FG-SBIR, which focused on early retrieval in SBIR, was first proposed in 2020, in which the author developed a novel reinforcement learning-based method (RL-based B) to optimize incomplete retrieval. Thus, based on previous studies, we choose the **RL-based B** method and some existing FG-SBIR baselines applied to the problem of on-the-fly FG-SBIR to verify the contribution of our proposed method. Five FG-SBIR baselines (B1, B2, B3, B4, and TS) mentioned in (Bhunia et al. 2020) were also selected for our experiments.

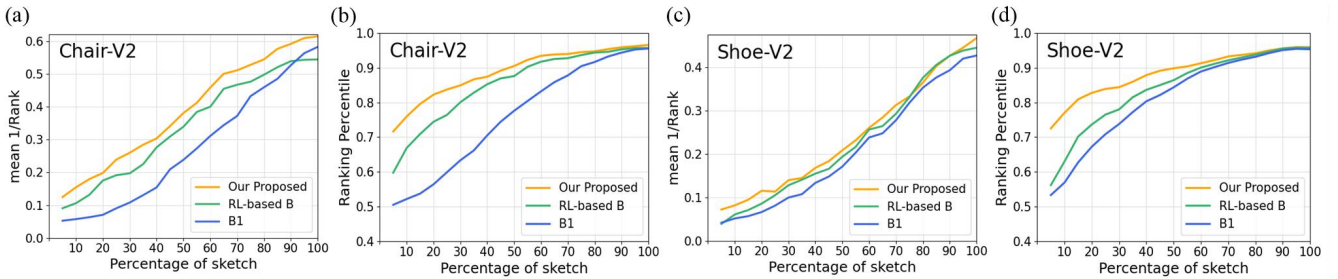

Figure 5. Performance of incomplete retrieval. Instead of showing the complete sketch (T = 20 sketch rendering steps), we visualize it using the percentage of sketch. A higher value indicates a better early retrieval performance.

## Performance Analysis

The performance of our proposed method (4-stage model) on the problem of on-the-fly SBIR is shown in **Fig.5** against the baseline methods (**RL-based B** and **B1**). We observe that (1) **B1** (state-of-the-art triplet loss based) performs poorly in the early retrieval, and the performance becomes better with the gradual completion of the incomplete sketch. This is because no mechanism in **B1** is designed to optimize the incomplete sketches or retrieval. (2) The performance of the **RL-based B** improved at the early retrieval owing to the optimization of the incomplete retrieval by reinforcement learning method; (3) The performance of our proposed method at early retrieval is significantly better against with that of the **B1** and **RL-based B**; Compared to the **B1**, our method designs the special modules to learn the efficient embedding space shared between the photo and its corresponding incomplete sketches. Moreover, considering the RL-based method, we optimize the embedding space of incomplete sketches by differentiable MGAL framework, instead of the non-differentiable ranking method. A qualitative result is shown in **Fig.5**, where RL-based B and B1 are the baselines.

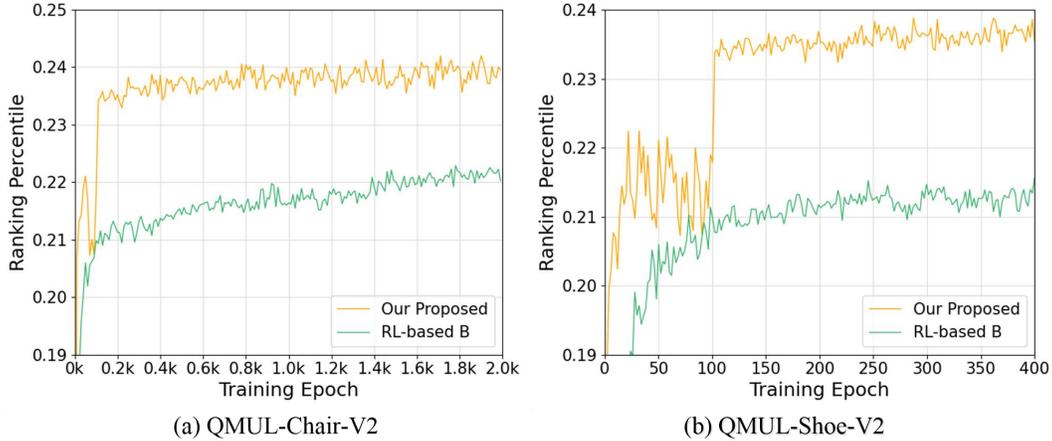

Figure 6. Training process of our proposed and RL-based B methods on QMUL-Chair-V2 and QMUL-Shoe-V2.

In addition to the five baselines (**B1, B2, B3, B4, TS, and RL-based B**), all the quantitative results are shown in **Table 1**; we can conclude that our proposed method outperforms all baselines by a significant margin, considering the early sketch retrieval performance, without deteriorating the top-5 and top-10 accuracy in the retrieval performance of the complete sketch. Moreover, owing to the design of a differentiable module to learn the embedding space for incomplete sketching, our proposed models converge faster than those of the RL-based B in the training process (see **Fig. 6**). We further evaluated the performance of our proposed with a varying feature embedding dimension. As shown in **Table 2**, we observe that our proposed method performs significantly better than that of the RL-based B at each dimension. The RL-based B performs best at 64-dimensional feature embeddings, whereas considering our method, larger dimensional vectors can also improve the performance appropriately.

**Table 1.** Comparative results with different baseline methods.

|  | Chair-V2 | | | | Shoe-V2 | | | |
|---|---|---|---|---|---|---|---|---|
|  | m@A | m@B | A@5 | A@10 | m@A | m@B | A@5 | A@10 |
| B1 | 77.18 | 29.04 | 76.47 | 88.13 | 80.12 | 18.05 | 65.69 | **79.69** |
| B2 | 80.46 | 28.07 | 74.31 | 86.69 | 79.72 | 18.75 | 61.79 | 76.64 |
| B3 | 76.99 | 30.27 | 76.47 | 88.13 | 80.13 | 18.46 | 65.69 | 79.69 |
| B4 | 81.24 | 29.85 | 75.14 | 87.69 | 81.02 | 19.50 | 62.34 | 77.24 |
| TS | 76.01 | 27.64 | 73.47 | 85.13 | 77.12 | 17.13 | 62.67 | 76.47 |
| RL-based B | 85.44 | 35.09 | 76.34 | 89.65 | 85.38 | 21.44 | **65.77** | 79.63 |
| Ours | **88.95** | **39.16** | **81.73** | **92.56** | **88.58** | **24.24** | 65.31 | 78.22 |

Table 2. Comparative results with varying feature-embedding.

| | Chair-V2 | | | | Shoe-V2 | | | |
|---|---|---|---|---|---|---|---|---|
| | m@A | | m@B | | m@A | | m@B | |
| D | RL-B | Ours | RL-B | Ours | RL-B | Ours | RL-B | Ours |
| 32 | 82.61 | 86.52 | 34.67 | 34.07 | 82.94 | 88.26 | 19.61 | 24.76 |
| 48 | - | 87.51 | - | 36.43 | - | **89.04** | - | 24.89 |
| 64 | 85.44 | 88.95 | 35.09 | **39.16** | 85.38 | 88.58 | 21.44 | 24.24 |
| 96 | - | 88.01 | - | 36.54 | - | 88.99 | - | **25.81** |
| 128 | 84.71 | **88.98** | 34.49 | 38.13 | 84.61 | 88.21 | 20.81 | 25.79 |

**Ablation Study**

Multiple incomplete sketches of one photo can confuse both the base triplet network and linear layer. Therefore, we proposed a multi-stage model to learn the embedding space for the complete sketch episode. We faced a problem when building such model. Determining the number of parts needed to divide a complete sketch episode was challenging. We test the performance of our models with varying stages, note that the one-stage model indicates that the complete sketch episode only learns one linear mapping function without association mechanism.

Table 3. Comparative results of our models with different stages and the RL-based B (RL-B) method.

| Models | Chair-V2 | | Shoe-V2 | |
|---|---|---|---|---|
| | m@A | m@B | m@A | m@B |
| RL-B (1-stage) | 85.44 | 35.09 | 85.38 | 21.44 |
| 1-stage | **89.18** | 38.77 | 87.77 | 21.97 |
| 2-stage | 89.08 | 39.01 | **88.64** | 23.68 |
| 3-stage | 88.95 | 38.97 | 88.55 | 24.02 |
| 4-stage | 88.95 | 39.16 | 88.58 | **24.24** |
| 5-stage | 88.88 | 39.11 | 88.39 | 24.05 |
| 6-stage | 88.74 | **39.35** | 88.48 | 24.08 |

The comparative results of the models with varying stages on the problem of the on-the-fly SBIR, as shown in **Table 3**. We make the following observations. (1) The performance of all our multi-stage models is significantly improved at early retrieval compared to the RL-based B method owing to our proposed approximation of the feature space of each incomplete sketch to that of both the target photo and its subsequent sketches. Even our one-stage model can still perform better than the RL-based B method (also one-stage). (2) Considering the increase in the stages, the performance of the models can become better because the phased learning of a complete sketch drawing episode can effectively mitigate the confusion in the model caused by the incomplete sketch diversity. Nonetheless, too many stages will also lead to over-fitting of the model and a decline in the performance.

**Further Analysis:**

Regarding practical applications, on-the-fly FG-SBIR still faces many challenges. One is that users' painting styles are usually different; therefore, sketch drawings reveal significant differences among users. For example, considering the same photo, different users would draw from different starting positions, and even considering the same stroke, they would draw differently (such as shape and size). These differences in strokes could lead to the difference for all the subsequent sketches, as shown in Fig.7. In the above experiments, each photo corresponds to only one complete sketch drawing episode. In this section, we generate sketch drawing episodes from different painting styles, that is, each photo can correspond to several sketch drawing episodes, and analyze the influence of the painting style on early retrieval results.

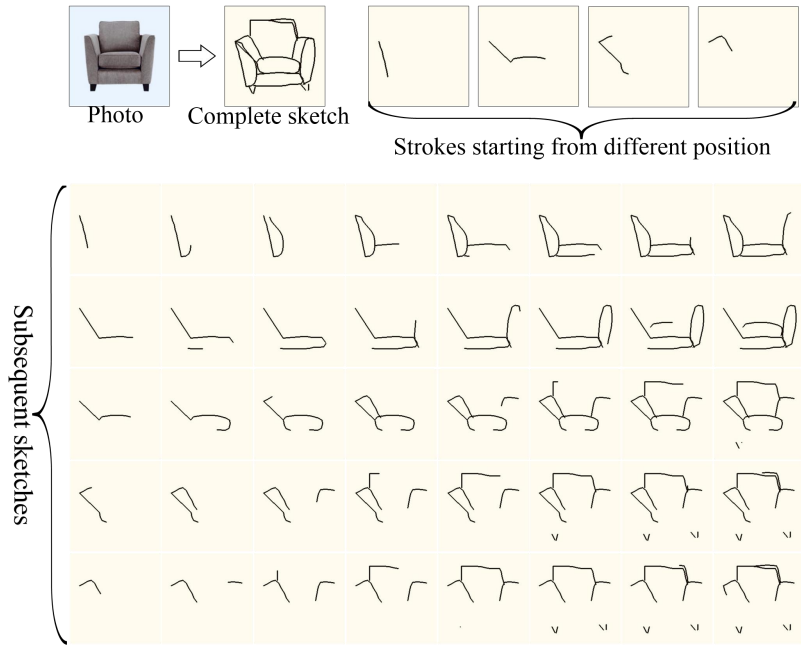

**Figure.7** An example for one photo corresponds to several sketch drawing episodes with different drawing styles.

**A. Different starting position for drawing a sketch**

For a photo, different users may draw each stroke starting from different positions of a photo (see **Fig.7**), which is normal. First, we discuss the influence of such differences on the performance of the above models. Each photo can generate one sketch drawing episode based on one default order of strokes (Order-0). Here, each photo generates four sketch drawing episodes based on four orders (Order-0, Order-1, Order-2, Order-3), other three orders was obtained by random modifying starting positions of Order-0. In the above sections, our proposed and the baselines were both trained on sketches generated using default order, and then were applied to the other three test sets (Orders-1, -2, and -3), respectively, as shown in **Fig. 8((a-0)~(a-3))**. We note that the order of strokes (one type of drawing style) can significantly affect the performance of the models.

In practice, the orders of strokes are highly diverse and far more than four. In order to verify the influence of such diversity on model learning, we designed three tasks (Task 1, Task 2, and Task 3). Task 1 indicates the original experiment in which each photo only generates one sketch drawing episode from Order-0. In Task 2, we extend the training and testing data by combining Orders-0 and -1; in Task 3, we extend the training and testing data by combining Orders-0, -1, -2, and -3. When training Tasks 2 and 3 using the same implementations as in Task 1. From **Fig. 8** and the results in **Table. 4,** we can observe that **(1)** the performance of models that trained on the augmented data improved (dashed column in **Fig. 8((b-0)~(b-3))**; **(2)** Different styles increase the diversity of data, and, in a moderate range, this diversity is equivalent to performing data augmentation; otherwise, the diversity will increase the difficulty of training the model. **(3)** When considering different painting styles, our proposed method outperforms the baseline by a significant margin.

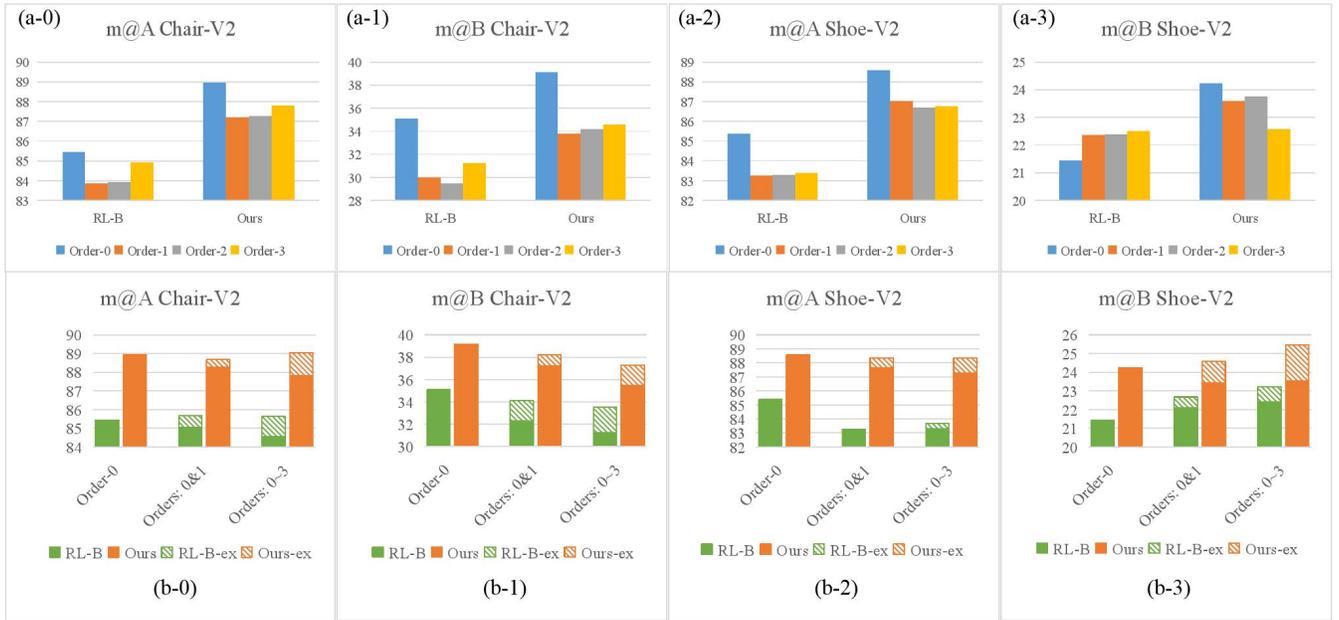

**Figure.8** The performance of our proposed and baseline on the sketch data with different drawing styles.

**Table.4** Comparative results on the extended dataset.

|  | Stroke | m@A | | m@B | | A@5 | | A@10 | |
|---|---|---|---|---|---|---|---|---|---|
|  |  | RL-B | Ours | RL-B | Ours | RL-B | Ours | RL-B | Ours |
| Chair-V2 | 1 | 85.44 | 88.95 | 35.09 | 39.16 | 76.34 | 81.73 | 89.65 | 92.56 |
| | 2 | 85.67 | 88.68 | 34.10 | 38.24 | 75.23 | 80.18 | 88.23 | 92.87 |
| | 4 | 85.63 | 89.05 | 33.53 | 37.30 | 75.23 | 82.04 | 88.85 | 91.02 |
| Shoe-V2 | 1 | 85.38 | 88.58 | 21.44 | 24.24 | 65.77 | 65.31 | 79.63 | 78.22 |
| | 2 | 83.20 | 88.36 | 22.67 | 24.58 | 62.91 | 64.41 | 76.42 | 77.47 |
| | 4 | 83.68 | 88.36 | 23.21 | 25.46 | 63.06 | 65.01 | 77.47 | 76.57 |

B. **Practical test**

Although one common painting style is considered, significant differences remain between users. In this section, we aim to verify the performance of our proposed and baseline models in practice and then expect to provide suggestions for future research. We invited 40 users (graduate students) to submit a total of 300 sketch drawing episodes (including 100 chair photos and 200 shoe photos) that we termed "users'" data. Before drawing, we showed the users the target photo and related several basic drawing rules. One instance drawn by different users is shown in **Fig. 9**, and, as shown in **Table 5,** the performances of the proposed and baselines were verified. The results confirm that (1) our proposed method notably outperformed the baseline methods (RL-based B) on the real hand-drawn sketches; (2) The performances of both the proposed and RL-based methods in practical testing were significantly improved compared with the test set, mainly because the strokes used to complete a sketch (users data) were far fewer than those used in model training (see **Fig. 9(b)**), that is, in practice, each stroke contained more details or information.

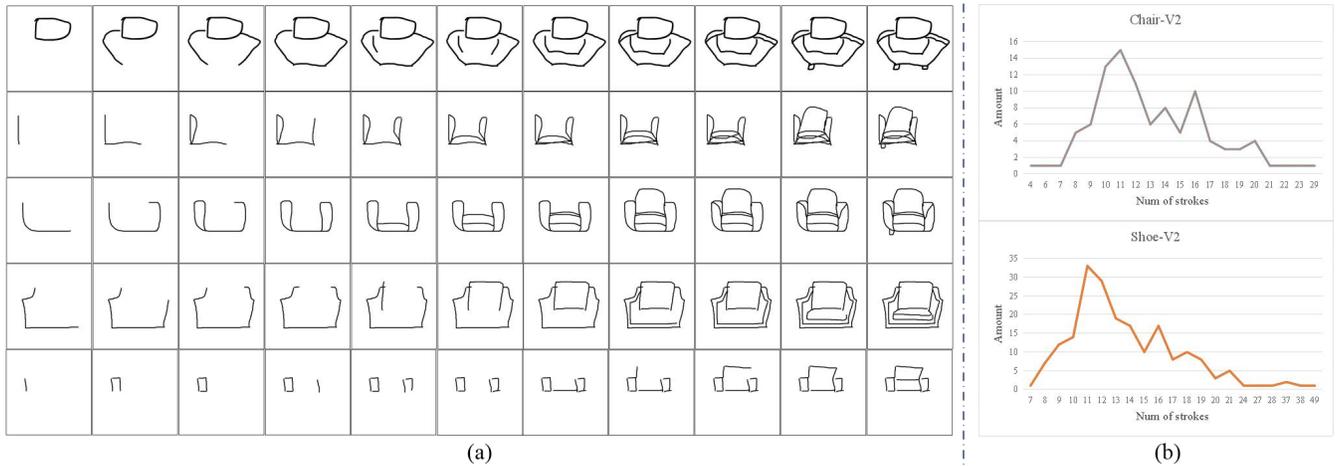

**Figure.9** Collecting real hand-drawn sketches from users. (a) One instance of several users drawing one photo; (b) Statistics of the stroke numbers of each complete sketch in the "users'" data.

Table.5 Comparative results on the users' data.

| | Source | Method | m@A | m@B | A@5 | A@10 |
|---|---|---|---|---|---|---|
| **Chair-V2** | Test set | RL-B | 85.44 | 35.09 | 76.34 | 89.65 |
| | | Ours | **88.95** | **39.16** | **81.73** | **92.56** |
| | Practice | RL-B | 88.28 | 34.70 | 71 | 86 |
| | | Ours | **91.28** | **39.05** | **80** | **91** |
| **Shoe-V2** | Test set | RL-B | 85.38 | 21.44 | 65.77 | 79.63 |
| | | Ours | **88.58** | **24.24** | 65.31 | 78.22 |
| | Practice | RL-B | 87.47 | 30.76 | 71.5 | 85.5 |
| | | Ours | **90.10** | **31.75** | 69 | 85 |

## Conclusion and Future Work

Recently, an increasing number of hand-painted data have appeared on the Internet, which makes FG-SBIR problems a topic of research in the field of computer vision. However, drawing a complete sketch often requires time and skills, which hinder its widespread adoption. A new problem known as the on-the-fly FG-SBIR was proposed to overcome the aforementioned barriers, where the retrieval was performed at every stroke drawn. We proposed a new framework of multi-granularity association learning to address on-the-fly FG-SBIR, in which the embedding space of an early incomplete sketch is guided to approximate both the target photo and its subsequent sketches. The experiments verify that our proposed method provides a considerable improvement.

In the practical applications, on-the-fly FG-SBIR still faces many challenges. Obviously, sketch modality can show greater diversity. For example, users may draw a sketch from different position of a object, users may draw the same sketch using different size, and even considering the same stroke, different users can draw differently. There are other factors that we have not taken into account. All of these realistic problem can affect the performance of the models obviously, which can reduce the generalization ability of the model and hinder its practical application. Such diversities should not be ignored in the future work. Moreover, we proposed a framework of MGAL, which was implement by a simple way; More advanced models should be used to implement our framework.